\documentclass[10pt,twocolumn,letterpaper]{article}

\usepackage[pagenumbers]{cvpr} 

\usepackage{graphicx}
\usepackage{amsmath}
\usepackage{mathtools} 
\usepackage{amssymb}
\usepackage{booktabs}
\usepackage{enumerate}
\usepackage{paralist}
\usepackage{xcolor}
\usepackage[accsupp]{axessibility}  

\usepackage[pagebackref,breaklinks,colorlinks]{hyperref}

\usepackage[capitalize]{cleveref}
\crefname{section}{Sec.}{Secs.}
\Crefname{section}{Section}{Sections}
\Crefname{table}{Table}{Tables}
\crefname{table}{Tab.}{Tabs.}

\def\cvprPaperID{2817} 
\def\confName{CVPR}
\def\confYear{2022}

\begin{document}
\title{Bridge-Prompt: Towards Ordinal Action Understanding in Instructional Videos}

\author{Muheng Li$^1$, Lei Chen$^1$, Yueqi Duan$^2$, Zhilan Hu$^3$, Jianjiang Feng$^1$, Jie Zhou$^1$, Jiwen Lu$^{\dag,1}$\\
$^1$Department of Automation, Tsinghua University\\
$^2$Department of Electronic Engineering, Tsinghua University\\
$^3$Media Technology Institute, Huawei Technologies Co., Ltd.}
 

\maketitle
\begin{abstract}
Action recognition models have shown a promising capability to classify human actions in short video clips. In a real scenario, multiple correlated human actions commonly occur in particular orders, forming semantically meaningful human activities. Conventional action recognition approaches focus on analyzing single actions. However, they fail to fully reason about the contextual relations between adjacent actions, which provide potential temporal logic for understanding long videos. In this paper, we propose a prompt-based framework, Bridge-Prompt (Br-Prompt), to model the semantics across adjacent actions, so that it simultaneously exploits both out-of-context and contextual information from a series of ordinal actions in instructional videos. More specifically, we reformulate the individual action labels as integrated text prompts for supervision, which bridge the gap between individual action semantics. The generated text prompts are paired with corresponding video clips, and together co-train the text encoder and the video encoder via a contrastive approach. The learned vision encoder has a stronger capability for ordinal-action-related downstream tasks, e.g. action segmentation and human activity recognition. We evaluate the performances of our approach on several video datasets: Georgia Tech Egocentric Activities (GTEA), 50Salads, and the Breakfast dataset. Br-Prompt achieves state-of-the-art on multiple benchmarks. Code is available at: \url{https://github.com/ttlmh/Bridge-Prompt}.

\end{abstract}

\let\thefootnote\relax\footnotetext{$^{\dag}$Corresponding author.}

\section{Introduction}

\begin{figure}
   \includegraphics[width=1\linewidth]{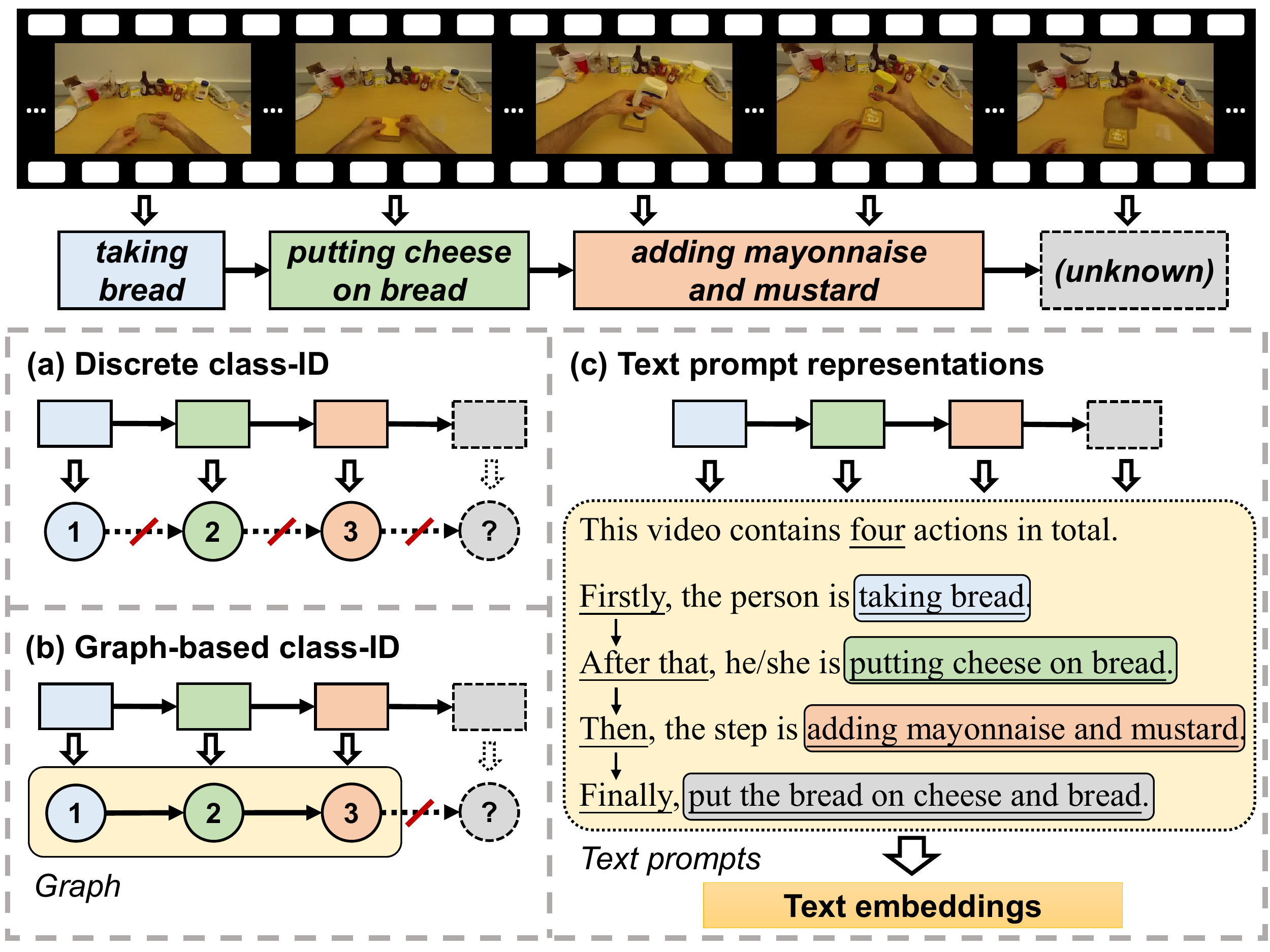}
    \caption{Comparisons of conventional representations and Bridge-Prompt representations for ordinal actions. The human activity of \textit{making cheese sandwich} contains four actions. Suppose the final action \textit{putting the bread on cheese and bread} is unseen in training set. Conventional approaches in (a) and (b) are unable to depict the intra-semantics and inter-relations of all four actions, while our Bridge-Prompt representations in (c) is able to capture the full semantic information.}
\label{fig:fig1}
\end{figure}

Recent years have witnessed the flourish of video analysis. Understanding human actions is the key to analyzing massive amounts of video data, which is conducive to a wide range of applications including video retrieval~\cite{Dzabraev_2021_CVPR}, video captioning~\cite{Luo2020UniVL} and video summarization~\cite{apostolidis2021video}. Among the many sub-topics in action analysis, action recognition is a basic and core issue, which has made remarkable progress under various well-designed models~\cite{carreira2017quo, feichtenhofer2019slowfast, Arnab_2021_ICCV}.

Meanwhile, the current research trend of video analysis is experiencing a transition from understanding single-semantics short video clips to longer and more complex videos~\cite{tang2019coin}. The increased attention on instructional video analysis has shown the significance of understanding semantically rich video contents~\cite{miech2019howto100m, 2019Cross,tang2019coin}. From the perspective of action analysis, conventional action recognition approaches focus on classifying the single action being performed in a short video clip~\cite{carreira2017quo, soomro2012ucf101}. In contrast, instructional video analysis methods need to study a series of actions being performed in longer time duration. In order to analyze instructional videos, we do not only need to understand the semantics of individual actions, but are also required to learn the semantic relations between contextual actions. 
Recently, some works have studied the mutual information between correlated actions in instructional videos using graph-based models~\cite{huang2020improving, zhou2021graph, nawhal2021activity}. The common approach is to regard each kind of action as a single node on a graph, where the edges between the nodes represent the contextual relations between adjacent actions. 

However, the graph-based approaches are transductive, which are limited by the prior knowledge of input nodes and/or edges. Therefore, graph-based approaches are unable to address unknown types of nodes and thus are hard to extend and transfer. Moreover, under the existing framework of action recognition, the current way of depicting human actions is to allocate individual annotations to every single action, where different actions are treated as separate class IDs. This is practicable for recognizing separate actions, yet it is unable to depict the contextual relations between ordinal actions since individual class IDs cannot provide contextual information. The example of (a) and (b) in Figure~\ref{fig:fig1} further illustrates the limitations of conventional class-ID-based approaches.

In this paper, we discover that human language is a powerful tool to depict the ordinal semantics between correlated actions. 
Human language is able to describe multiple sequentially occurred events based on ordinal numerals and specific sentence patterns. For example, ordinal relations between \textit{taking bottle} and \textit{pouring water} can be described in:``\textit{the person firstly takes (the) bottle, and then pours water (into it)}''. The language naturally bridges the semantics between ordinal actions. In certain circumstances, even the textual descriptions of actions themselves can provide contextual information. For example, the ordinal relationship between actions of \textit{taking bread}, \textit{putting cheese on bread} and \textit{putting bread on cheese and bread} is easy to be deduced literally. Moreover, language can intuitively extrapolate to unknown types of action. Given a new expression \textit{putting bread on bread}, its semantics can be inferred from the expressions of known types of action. Figure~\ref{fig:fig1}(c) illustrates the effectiveness of language representations.

To this end, we propose a text-based learning method, Bridge-Prompt, for instructional video analysis. Motivated by the recent advances of prompt-based learning approach in Natural Language Processing (NLP)~\cite{liu2021pre} and visual recognition~\cite{CLIP2021}, we introduce a three-plus-one-level design of text prompts to analyze the video clips containing a series of ordinal actions. Figure~\ref{fig:fig1} shows the comparisons between conventional and Bridge-Prompt representations of ordinal actions. More specifically, we develop a prompt-based learning framework to jointly co-train the video and text encoders based on a specially designed video-text fusion module, so that we simultaneously exploit out-of-context and contextual action information towards a more comprehensive understanding of instructional videos. Our work digs deeper into the further potential of prompt-based learning approaches towards ordinal action understanding and instructional video analysis. Extensive experimental results on three benchmark datasets illustrate that the Bridge-Prompt-based approaches have achieved promising performances, and reach state-of-the-art on several benchmarks with the help of the prompt-based learning framework.   




\begin{figure*}
\begin{center}
\includegraphics[width = 1\linewidth]{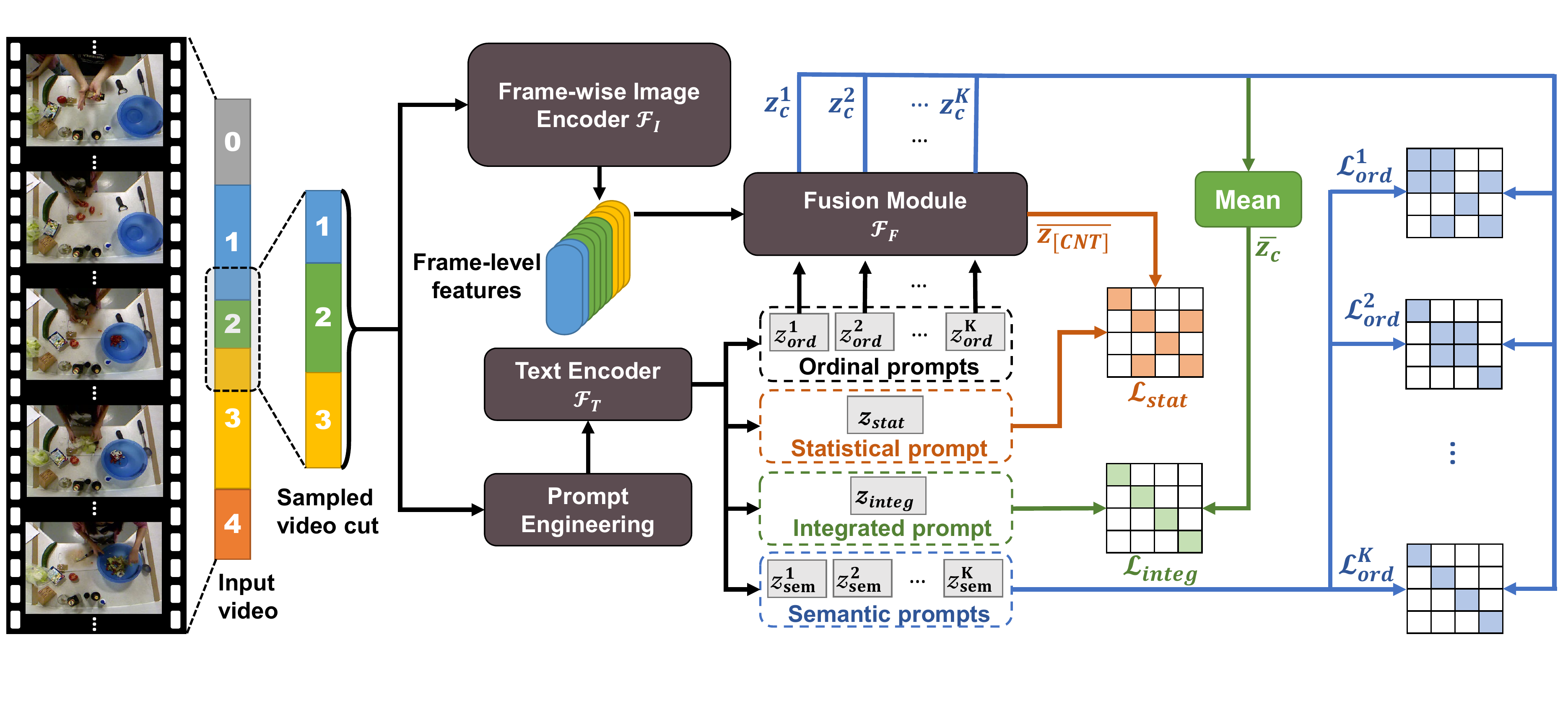}
\end{center}
\vspace{-20pt}
\caption{Overview of Bridge-Prompt pipeline. Bridge-Prompt takes the video cuts from minute-long raw inputs. After the special prompt engineering procedure, four types of text prompts are generated. Vision and text information are integrated both in the fusion module and during the video-text contrastive learning process. The proposed pipeline is able to capture the relations between ordinal actions.}
\label{fig:pipeline}
\vspace{-5pt}
\end{figure*}

\section{Related Work}

\noindent\textbf{Action analysis on instructional videos.} Instructional video analysis is an increasingly popular trend in the field of video understanding. A wide variety of instructional video datasets have been proposed in recent years~\cite{tang2019coin, zhukov2019cross, miech2019howto100m, zhou2018towards }. Instructional videos include profuse semantic information of human activities. The conventional approaches on action recognition~\cite{2014stream, 3dcnn, feichtenhofer2019slowfast, liu2021video} mainly focus on the datasets of trimmed video clips containing a single action in each video clip~\cite{soomro2012ucf101, carreira2017quo}. Based on the existing studies of action recognition, several works have extended the action analysis methods to instructional videos by paying attention to the relations between ordinal actions. GTRM~\cite{huang2020improving} utilizes a graph-based structure to depict the ordinal actions, and analysis is based on Graph Convolutional Networks (GCNs)~\cite{graph2017}. GHRM~\cite{zhou2021graph} also represents ordinal actions as a graph, while focusing on the long-term action recognition task. Besides, Shao \textit{et al.}~\cite{shao2020intra} proposed the TransParser method for intra- and inter-action understanding via temporal action parsing. Different from the previous solutions, we make use of human language as a powerful semantic tool for analyzing ordinal actions in instructional videos. 

\noindent\textbf{Prompt-based learning on computer vision.} Prompt-based learning approaches have been extensively studied in NLP~\cite{PET2021, autoprompt:emnlp20, liu2021pre}. The pioneer language model as GPT-3~\cite{NEURIPS2020_1457c0d6} has shown its great few-show or zero-shot potential across various tasks. The core of prompt-based learning is to modify the input sample as a prompted version and embed the expected output information as an unfilled slot inside the prompt. CLIP~\cite{CLIP2021} introduces the prompt-based learning approach into the image recognition task by embedding the textual labels of the to-be-recognized objects into descriptive texts, and the classification procedure can be transformed into a video-text matching problem. Following the prompt-based design, ALIGN~\cite{align2021} scales up the vision-language model by training on over one billion noisy image-text pairs and achieves better prompt-based prediction performances than CLIP. CoOp~\cite{zhou2021coop} utilizes learnable tokens as textual prompts and gains a promotion on few-shot image classification. CLIP-Adapter~\cite{gao2021clip} combines the adapted features generated by the designed feature adapter with the CLIP feature to fit the few-shot classification. The prompt-based learning approach has not been widely developed on video understanding. ActionCLIP~\cite{wang2021actionclip} proposes a specially designed prompt-based paradigm for action recognition, but it mainly focuses on recognizing single actions in short video clips. Our proposed Bridge-Prompt aims at analyzing instructional videos, which is more challenging but more conducive to understanding human behaviors.

\section{Method}
In this section, we introduce the overall pipeline design of Bridge-Prompt. The pipeline of our approach is illustrated in Figure~\ref{fig:pipeline}. 

\subsection{Prompt Engineering}
Prompt engineering refers to the design of an input text template that embeds the expected output strings as fill-in-the-blank formats~\cite{NEURIPS2020_1457c0d6} (\eg, cloze test). The objective of our prompt engineering procedure is to design specific forms of text prompts to describe groups of ordinal actions in instructional videos. Suppose a series of single actions ($A=\{a_1,a_2,...,a_K\}$) composes a specific kind of human activity. An easier way to design the prompts is to pose a blank-filling problem for every single action. For example, the prompt format as ``\textit{the person is \underline{\{vp${}_i$\}} right now}" (\textit{vp${}_i$} refers to the verb-phrase description for action $a_i$) can be used to abstract the semantics for each separate action of the character. However, since each action is still treated as an independent prompt instance, this strategy is unable to depict the contextual semantics between adjacent ordinal actions. For example, within the human activity of \textit{scrambling egg}, the \textit{stir-frying egg} action can only happen after \textit{cracking egg}. A better form of text prompts towards ordinal action analysis should not only capture the out-of-context semantics of each separate action, but also bridge the gap between contextually related actions, and depict the overall semantics of the series of actions.

To better represent the series of actions in the Bridge-Prompt framework, we propose a three-plus-one-level design of prompt engineering for instructional videos: statistical prompt, ordinal prompt, semantic prompt, and integrated prompt. Considering the input video cut with $K$ consecutive actions: 

\noindent\textbf{1) Statistical prompt} captures the total count information for the series of actions. We use the format as ``\textit{this video clip contains \underline{\{num(K)\}} actions in total}". The statistical prompt is denoted as $y_{stat}$.

\noindent\textbf{2) Ordinal prompt} captures the positional information for each action. We use the format as ``\textit{this is the \underline{\{ord$_i$\}} action in the video}". The ordinal prompt is denoted as $y^i_{ord}$. The ordinal prompt set for $x$ is denoted as:
\begin{equation}
    \mathcal{Y}_{ord}=[y^1_{ord},...,y^K_{ord}]
\end{equation}
\noindent\textbf{3) Semantic prompt} is the core of prompt design, which captures the semantic information of the actions. To integrate both out-of-context and contextual action information, we merge the ordinal information into the semantic prompts to create a multi-prompt format. We use the format as ``\textit{\underline{\{ord$_i$\}}, the person is performing the action step of \underline{\{vp${}_i$\}}}" for action $a_i$. The semantic prompt set for $x$ can be denoted as:
\begin{equation}
    \mathcal{Y}_{sem}=[y^1_{sem},...,y^K_{sem}]
\end{equation}
\noindent\textbf{3+1) Integrated prompt} captures the overall information for video $x$. The integrated prompt is formed by the integration of all semantic prompts $\mathcal{Y}_{sem}$. The integrated prompt $y_{integ}$ can be denoted by:
\begin{equation}
    y_{integ}=y^1_{sem}\oplus y^2_{sem}\oplus...\oplus y^K_{sem}
\end{equation}
where $\oplus$ refers to the string concatenation operation.


{
\small
\begin{figure}
   \includegraphics[width=1\linewidth]
                   {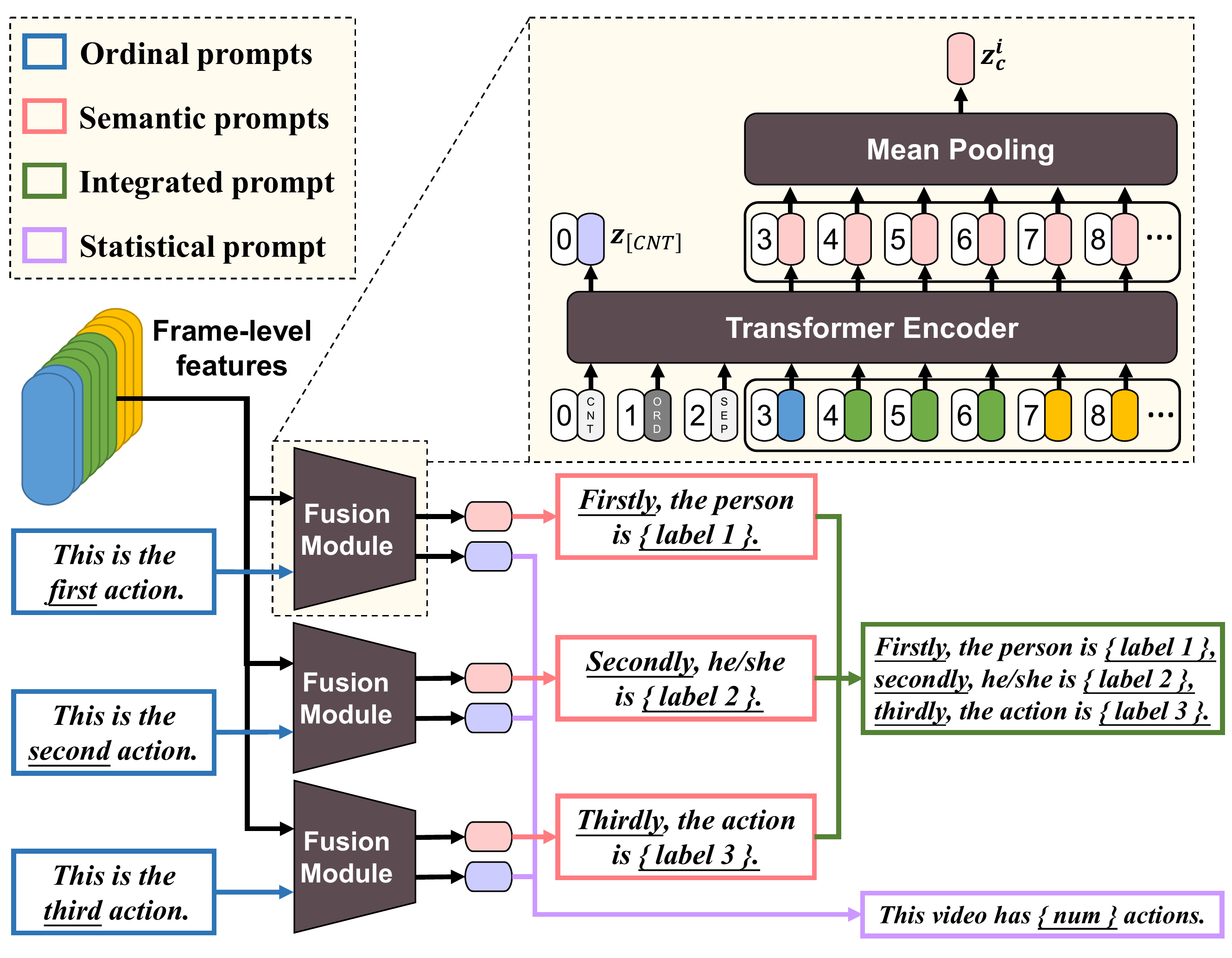}
   \caption{Detailed illustration of prompt formats and fusion encoder. The fusion encoder takes the encoded frame-wise features and the ordinal prompt embeddings as inputs. It employs a learnable count token to analyze the statistical information. We adopt an ordinal-attention manner, meaning that the module only focuses on a single action with respect to a particular ordinal each time. The integrated semantics is extracted by mean-pooling operation.}
\label{fig:snce}
\vspace{-5pt}
\end{figure}
}

\subsection{Bridge-Prompt: Framework}

\noindent\textbf{Sampling for raw videos. }
The raw instructional video sample $x_0\in\mathbb{R}^{L_0\times 3\times H\times W}$ contains $L_0$ RGB frames of size $H\times W$. Usually, $L_0$ is different for each raw video. Moreover, suppose $K_0$ actions are contained in $x_0$, and $K_0$ is also unequal for different activities. Within each video, the duration of each action is unevenly distributed. We propose a sampling strategy by generating random video cut $x\in\mathbb{R}^{L_c\times 3\times H\times W}$ from raw videos of a fixed length $L_c$ to extract useful information while improving training efficiency. Each cut $x$ may contain a single action or several successive actions, where $K$ denotes the action count for $x$. The prompt engineering is conducted on those video cuts to generate the corresponding prompted text pair $y$. The sampling operation actually limits the temporal reception field of the model to a more localized range. The advantage of such a sampling strategy is to force the Bridge-Prompt model to focus more on the logical connections both within and between locally related actions.

\noindent\textbf{Pre-training pipeline. } The sampled video cut $x$ with $L_c$ frames $[f_1,...,f_{L_c}]$ firstly passes through a frame-wise image encoder $\mathcal{F}_{I}$ to generate the frame-level features $[\mathcal{F}_{I}(f_1),...,\mathcal{F}_{I}(f_{L_c})]$. Meanwhile, according to the prompt rules, a set of textual prompts $\{y_{stat}, \mathcal{Y}_{ord}, \mathcal{Y}_{sem}, y_{integ}\}$ can be generated for $x$. A text encoder $\mathcal{F}_{T}$ is introduced to extract the textual prompt embeddings  $\{\textbf{z}_{stat}, \mathcal{\textbf{Z}}_{ord}, \mathcal{\textbf{Z}}_{sem}, \textbf{z}_{integ}\}$ respectively. The frame-level features are then passed through a fusion encoder $\mathcal{F}_{F}$ together with ordinal prompt embeddings to extract the clip-level feature $\textbf{z}^i_{c}=\mathcal{F}_{F}(\mathcal{F}_{I}(f_1),...,\mathcal{F}_{I}(f_{L_c}), \textbf{z}^i_{ord})$ for the i-th action of $x$. The design for the fusion module is the key to understanding both intra-action and inter-action information in $x$. We propose a Transformer-based structure for fusion. The information of ordinal prompt $y^i_{ord}$ is fused into the fusion encoder to provide instructive information. We also embed a count token inside $\mathcal{F}_{F}$ to collect the quantitative information to be matched with statistical prompt $y_{stat}$. The details of the fusion approach for Bridge-Prompt pre-training will be discussed in the following sub-section. The clip-level feature is jointly learnt with both semantic prompts $\mathcal{Y}_{sem}$ and integrated prompt $y_{integ}$ under a contrastive vision-text learning pattern.

\noindent\textbf{Fusion module. } The fusion encoder extracts the core information from the consecutive frame-level features. In other words, it tries to abstract the series of actions that occur in the input video clip. We utilize an ordinal-attention manner for the fusion module, \ie, each time the fusion module only focuses on the action of a specific location. The ordinal-attention mechanism is implemented by adding the i-th ordinal prompt embeddings $\textbf{z}^i_{ord}$ to the fusing inputs, which is an early-fusion strategy. We utilize a Transformer-Encoder structure for the fusion module. The input tokens of the fusion encoder include a learnable count token \texttt{[CNT]}, $\textbf{z}^i_{ord}$ as a token \texttt{[ORD]}, a split token \texttt{[SEP]}, and $L_c$ visual tokens representing frame-level features. \texttt{[ORD]} indicates which number of actions the fusion encoder is focusing on. The encoded representations of $L_c$ frame-level features are mean-pooled to represent the clip-level feature. Besides, we added a learnable count token to learn additional quantitative information of actions. The encoded representation $\textbf{z}_{[CNT]}$ for \texttt{[CNT]} will pass through the same contrastive vision-text learning framework with statistical prompt embeddings $\textbf{z}_{stat}$ as a clip-level feature.

\noindent\textbf{Joint vision-text representation learning. } The joint vision-text representation learning maximizes the similarity between the encoded vision features and text features. A video clip $x$ and its text description $y$ can be encoded respectively with a video encoder and a text encoder, generating the clip representation $\textbf{z}_x$ and the text representation $\textbf{z}_y$. The similarity between $\textbf{z}_x$ and $\textbf{z}_y$ can be defined as their cosine distance:
\begin{equation}
    s(\textbf{z}_x, \textbf{z}_y)=\frac{\textbf{z}_x\cdot\textbf{z}_y}{\left\vert\textbf{z}_x\right\vert\left\vert\textbf{z}_y\right\vert}
\end{equation}
For a batch of the clip features $\mathcal{Z}_x$ and its corresponding batch of text features $\mathcal{Z}_y$, the batch similarity matrix $S$ is:
\begin{equation}
S(\mathcal{Z}_x,\mathcal{Z}_y)=
\begin{bmatrix}
s(\textbf{z}_{x_1}, \textbf{z}_{y_1})  & \cdots   & s(\textbf{z}_{x_1}, \textbf{z}_{y_B})   \\
\vdots  & \ddots   & \vdots  \\
s(\textbf{z}_{x_B}, \textbf{z}_{y_1})  & \cdots\  & s(\textbf{z}_{x_B}, \textbf{z}_{y_B})  \\
\end{bmatrix}
\end{equation}
where $B$ is the batch size. A text-wise/clip-wise softmax-normalization function can be applied respectively along rows/columns on $S(\mathcal{Z}_x,\mathcal{Z}_y)$, generating $S_T(\mathcal{Z}_x,\mathcal{Z}_y)$ and $S_V(\mathcal{Z}_x,\mathcal{Z}_y)$. A ground truth batch similarity matrix $GT$ is defined where the similarity score of positive pair equals to $1$, while negative pair equals 0. Our objective is to maximize the similarity between $S$ and $GT$. We define the Kullback–Leibler (KL) divergence for matrices as the multi-modal contrastive loss:
\begin{equation}
    D_{KL}(P\Vert Q)=\frac{1}{N^2}\sum_{i=1}^{N}{\sum_{i=1}^{N}{P_{ij}\log{\frac{P_{ij}}{Q_{ij}}}}}
\end{equation}
where $P$ and $Q$ are $N\times N$ matrices. The contrastive loss for video-text pair can be defined as:
\begin{equation}
    \mathcal{L}=\frac{1}{2}\left[D_{KL}(S_T\Vert GT)+D_{KL}(S_V\Vert GT)\right]
\end{equation}
Under the Bridge-Prompt framework, there are three parts of video-text contrastive losses in total:
\begin{enumerate}[i)]
	\item $\textbf{z}^i_c$ fused by the i-th ordinal token with $\textbf{z}^i_{sem}$ of corresponding ordinal prompt, notated as $\mathcal{L}^i_{sem}$;
	\item mean-pooled $\overline{\textbf{z}_c}$ fused by all ordinal tokens with $\textbf{z}_{integ}$, notated as $\mathcal{L}_{integ}$;
	\item mean-pooled $\overline{\textbf{z}_{[CNT]}}$ with $\textbf{z}_{stat}$, notated as $\mathcal{L}_{stat}$;
\end{enumerate}
The overall loss objective for the Bridge-Prompt pre-training framework is as follows:
\begin{equation}
    \mathcal{L}=\sum^{K}_{i=1}{\mathcal{L}^i_{sem}}+\lambda_1\mathcal{L}_{integ}+\lambda_2\mathcal{L}_{stat}
\end{equation}
where $\lambda_1$ and $\lambda_2$ balance the three losses.

\subsection{Prompt-Based Inference}
The \textit{``pre-train, prompt, and predict''} paradigm in NLP has suggested that prompt-based design has the superiority of combining the objectives of downstream tasks into the pre-training procedure. The Bridge-Prompt framework has the capability of recognizing a series of actions by solving prompt-based cloze tests as ``\textit{this video clip contains \underline{\ \ \ \ } actions in total}" or ``\underline{\ \ \ \ }, \textit{the person is performing the action of \underline{\ \ \ \ }}". 
In practice, we first generate the text features for all relevant ordinal prompts, statistical prompts, and semantic prompts by the pre-trained text encoder. For each test video, we extract the clip-wise features embedded by different ordinal prompts $\textbf{z}^{i}_c$ and the average statistical representation $\overline{\textbf{z}_{[CNT]}}$ using the pre-trained image encoder and fusion encoder. At first, we find the most matched embedding of statistical prompts with $\overline{\textbf{z}_{[CNT]}}$ to determine the total count of actions. Then, we find the most matched embedding of semantic prompt with each ordinal prompt-embedded clip-wise feature $\textbf{z}^{i}_c$ to determine each ordinal action one by one. As for the prompt variants, we vote among all variant formats to get the most matched prompt during inference stage.

\section{Experiments}

\subsection{Datasets}

We evaluate our proposed model on three challenging datasets. \textbf{50Salads}~\cite{stein2013combining} contains 50 top-view 30-fps instructional videos regarding salad preparation. Totally 19 kinds of actions are contained in all videos. The 5-fold cross-validation is performed for evaluation, and the average results are reported. \textbf{Georgia Tech Egocentric
Activities (GTEA)}~\cite{fathi2011learning} contains 28 egocentric 15-fps instructional videos of daily kitchen activities. Totally 74 classes of actions are summarized from all videos. We use the 4-fold cross-validation to evaluate the performances, and the average results are reported. \textbf{Breakfast}~\cite{breakfast2014} contains 1,712 third-person 15-fps videos of breakfast preparation activities. 48 types of different actions are included in all 10 different kinds of breakfast activities. For evaluation, we use the train-split setting as proposed in~\cite{hussein2019timeception}, with 1357 videos for training and 355 videos for testing.

\subsection{Implementation Details}

\noindent\textbf{Sampling strategy. } The video cut sampling strategy is adjusted concerning frame rates and scales of different datasets. In general, we adopt a 16-frame window for each video cut. For GTEA dataset, we adopt multiple downsampling rates as 1, 2 and 4 respectively corresponding to the window striding rates of 2, 1 and 0.5. For 50Salads dataset, we use higher 24 and 32 downsampling rates with window striding rate of 1. For the Breakfast dataset, we a employ downsampling rate of 16 with a window striding rate of 2.

\noindent\textbf{Bridge-Prompt architectures. } For the image and text encoders, we follow the setups as CLIP~\cite{CLIP2021} and ActionCLIP~\cite{wang2021actionclip}. We adopt ViT-B/16~\cite{vit2021} as the image encoder $\mathcal{F}_I$, which is a 12-layer Transformer with input patch sizes of 16. The output representation for \texttt{[CLS]} token is regarded as the image feature. The text encoder $\mathcal{F}_T$ is also a 12-layer Transformer with the width of 512 and 8 attention heads. The output representation for \texttt{[EOS]} token is regarded as the text feature. The output frame-wise feature of the image encoder is a 768-dimensional vector, which is mapped to a 512-dimensional latent vector to match the embedded text features. For the fusion module $\mathcal{F}_F$, we employ a Transformer-Encoder-based structure to fuse the information of both image and text features. The fusion module contains 6 layers. As for the details of the prompt engineering procedure, we utilize an invariant prompt format for ordinal prompts and statistical prompts. With respect to the semantical prompts (which also contribute to integrated prompts), we adopt 19 variant prompt formats (9 short variant versions for integrated prompts) to describe the action semantics. The average similarity of all variants are computed during the prompt-based inference stage.

\noindent\textbf{Training details. } The image encoder and text encoder are together pre-trained on Kinetics-400\cite{carreira2017quo} by~\cite{wang2021actionclip} before our training. We adopt AdamW~\cite{AdamW} optimizer with the base learning rate of $5\times10^{-6}$ with a 0.2 weight decay. The first 10\% of training epochs are set as a warm-up phase, and the learning rate gradually decays down to zero during the remaining epochs under a cosine schedule. The spatial resolution of the input video is $224\times224$. For the loss function, we simply set $\lambda_1=\lambda_2=1$. The model is trained for 50 epochs on GTEA and 50Salads, and 35 epochs on Breakfast. We use the batch size of 12 during training.

\subsection{Results on Action Segmentation}

The objective of action segmentation is to classify the action that occurs in each frame of a video~\cite{lea2017temporal}. Different from action recognition, action segmentation processes videos with multiple action instances. In consequence, action segmentation approaches should not only understand the out-of-context semantics for each separate action, but also be aware of the logical relations between adjacent actions. Several works have been conducted, and have achieved promising segmentation results. Most of the current SOTA approaches on action segmentation utilize the frame-wise I3D~\cite{carreira2017quo} features pre-trained on Kinetics extracted by~\cite{farha2019ms}, since the videos used for action segmentation are generally long videos that are hard to conduct direct analysis based on raw data. Bridge-Prompt utilizes a video cut-based approach to learn the contextual relations between adjacent actions locally, which is feasible on long videos. Since our approach is not specially designed for end-to-end action segmentation, we mainly adopt the Bridge-Prompt pre-trained image encoders to generate frame-wise features for raw videos. We test the action segmentation results based on current segmentation backbones.

\noindent\textbf{Evaluation metrics. } To evaluate the action segmentation results, we adopt several metrics including frame-wise accuracy (Acc), segmental edit distance, and the segmental F1 score at overlapping thresholds \{10\%, 25\%, 50\%\}, denote by F1@\{10,25,50\}. The frame-wise accuracy is the most direct and frequently used metric, whereas it is unable to penalize the over-segmentation errors in long-duration actions. The segmental edit distance and segmental F1 score~\cite{lea2017temporal, lea2016segmental} are proposed to handle over-segmentation errors and measure the segmentation quality.

\noindent\textbf{Comparisons with the state-of-the-art. } We compare the segmentation performances based on Bridge-Prompt-encoded frame-wise features with previous state-of-the-art methods. We use the ASFormer~\cite{yi2021asformer} as the backbone model for proceeding action segmentation. Bridge-Prompt is used as the pre-training approach to train the frame-wise image encoder (ViT). The output 768-dimensional frame-wise representations are regarded as the training inputs for the action segmentation backbone. In comparison, the previous state-of-the-art approaches use 2048-dimensional I3D features as training inputs. We conduct action segmentation on the GTEA dataset and the 50Salads dataset.

\begin{figure*}
\begin{center}
\includegraphics[width = 1\linewidth]{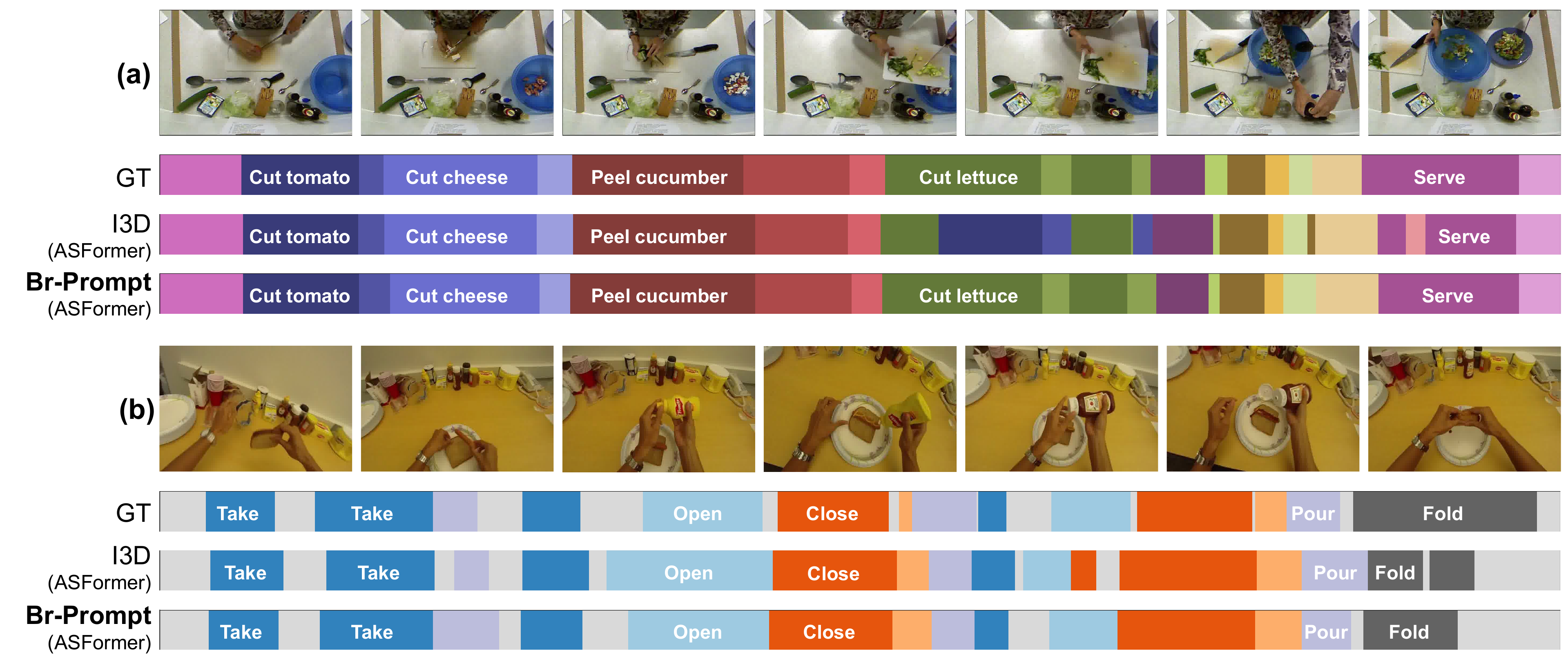}
\end{center}
\vspace{-10pt}
\caption{Qualitative results for action segmentation task on (a) 50Salads, and (b) GTEA dataset. Part of the actions are annotated on the color bar. The Br-Prompt pre-trained representation has greater potential on action segmentation task.}
\vspace{-10pt}
\label{fig:seg}
\end{figure*}

Table~\ref{table:gtea1},~\ref{table:salad1} compare the quantitative results of our approach. Specifically, we predict the 11 verbs of actions in GTEA for fair comparisons, and our method outperforms current state-of-the-art approaches under all five evaluation metrics. For comparison, we also evaluate the performances using raw features of ViT pre-trained by~\cite{wang2021actionclip}, which are inferior to the results using I3D-pre-trained features. However, after the ViT image encoder is further trained by Bridge-Prompt, the performances get obvious boosts. The performance of our approach also precedes previous state-of-the-art results on 50Salads. Figure~\ref{fig:seg} shows the qualitative illustration of action segmentation on both datasets.

\begin{table}[]
\small
\begin{center}
\setlength{\tabcolsep}{5pt}
\caption{Action segmentation results on GTEA dataset.}
\vspace{-5pt}
\begin{tabular}{l|ccc|c|c}
\toprule
\textbf{GTEA} & \multicolumn{3}{c|}{\textbf{F1}@\{10,25,50\}} & \textbf{Edit} & \textbf{Acc} \\ \hline
BCN~\cite{wang2020boundary} & 88.5 & 87.1 & 77.3 & 84.4 & 79.8 \\
MS-TCN++~\cite{li2020ms} & 88.8 & 85.7 & 76.0 & 83.5 & 80.1 \\
ASRF~\cite{ishikawa2021alleviating} & 89.4 & 87.8 & 79.8 & 83.7 & 77.3 \\
G2L~\cite{gao2021global2local} & 89.9 & 87.3 & 75.8 & 84.6 & 78.5 \\
SSTDA~\cite{chen2020action} & 90.0 & 89.1 & 78.0 & 86.2 & 79.8 \\
SSTDA+HASR~\cite{ahn2021refining} & 90.9 & 88.6 & 76.4 & 87.5 & 78.7 \\
ASFormer (I3D)~\cite{yi2021asformer} & 90.1 & 88.8 & 79.2 & 84.6 & 79.7 \\ \hline
ASFormer (ViT) & 88.5 & 86.2 & 77.6 & 87.1 & 75.6 \\
\textbf{Br-Prompt}+ASFormer & \textbf{94.1} & \textbf{92.0} & \textbf{83.0} & \textbf{91.6} & \textbf{81.2} \\
\bottomrule
\end{tabular}\label{table:gtea1}
\vspace{-15pt}
\end{center}
\end{table}

\begin{table}[]
\small
\begin{center}
\setlength{\tabcolsep}{5pt}
\caption{Action segmentation results on 50Salads dataset.}
\vspace{-5pt}
\begin{tabular}{l|ccc|c|c}
\toprule
\textbf{50Salads} & \multicolumn{3}{c|}{\textbf{F1}@\{10,25,50\}} & \textbf{Edit} & \textbf{Acc} \\ \hline
MS-TCN++~\cite{li2020ms} & 80.7 & 78.5 & 70.1 & 74.3 & 83.7 \\
BCN~\cite{wang2020boundary} & 82.3 & 81.3 & 74.0 & 74.3 & 84.4 \\
SSTDA~\cite{chen2020action} & 83.0 & 81.5 & 73.8 & 75.8 & 83.2 \\
ASRF~\cite{ishikawa2021alleviating} & 84.9 & 83.5 & 77.3 & 79.3 & 84.5 \\
ASFormer (I3D)~\cite{yi2021asformer} & 85.1 & 83.4 & 76.0 & 79.6 & 85.6 \\
ASFormer+ASRF (I3D) & \multicolumn{1}{l}{85.1} & \multicolumn{1}{l}{85.4} & \multicolumn{1}{l|}{79.3} & \multicolumn{1}{l|}{81.9} & \multicolumn{1}{l}{85.9} \\
SSTDA+HASR~\cite{ahn2021refining} & 86.6 & 85.7 & 78.5 & 81.0 & 83.9 \\ \hline
\textbf{Br-Prompt}+ASFormer & \textbf{89.2} & \textbf{87.8} & \textbf{81.3} & \textbf{83.8} & \textbf{88.1} \\
\bottomrule
\end{tabular}\label{table:salad1}
\vspace{-18pt}
\end{center}
\end{table}

\subsection{Results on Long-Term Activity Recognition}



A series of ordinal actions in instructional videos generally form a high-level semantics of human activity. The objective of long-term activity recognition is to classify the types of activities in long videos. Recognizing a high-level activity requires understanding the basic relations and temporal evolution of its ordinal sub-actions. Since Bridge-Prompt aims to study the relations between ordinal actions, it is also capable of long-term activity recognition. To adapt our framework for long-term action recognition, we first pre-train the frame-level encoder based on the Bridge-Prompt framework, and extract the frame-wise features for each video. Then, we uniformly sample 64 segments in each video with 8 frames per segment as in~\cite{hussein2019timeception}. We use a simple Transformer-Encoder as a fusion module to respectively integrate segment-wise frames and different segments to generate video-wise representations. Then the human activities are predicted using prompt-based inferences.

\begin{table}[]
\small
\begin{center}
\setlength{\tabcolsep}{24pt}
\caption{Human activity recognition results on Breakfast dataset.}
\vspace{-5pt}
\begin{tabular}{l|c}
\toprule
\multicolumn{1}{l|}{\textbf{Method}} & \textbf{Acc(\%)} \\ \hline
\multicolumn{2}{l}{Kinetics pre-trained I3D} \\ \hline
\multicolumn{1}{l|}{I3D~\cite{carreira2017quo}} & 58.61 \\
\multicolumn{1}{l|}{ActionVLAD~\cite{girdhar2017actionvlad}} & 65.48 \\
\multicolumn{1}{l|}{Timeception~\cite{hussein2019timeception}} & 67.07 \\
\multicolumn{1}{l|}{VideoGraph~\cite{hussein2019videograph}} & 69.45 \\
\multicolumn{1}{l|}{GHRM~\cite{zhou2021graph}} & 75.49 \\ \hline
\multicolumn{2}{l}{Breakfast fine-tuned} \\ \hline
\multicolumn{1}{l|}{I3D (fine-tuned)~\cite{zhou2021graph}} & 74.83 \\
\multicolumn{1}{l|}{\textbf{Br-Prompt} (fine-tuned)} & \textbf{80.00} \\
\bottomrule
\end{tabular}\label{table:acti}
\vspace{-20pt}
\end{center}
\end{table}

\noindent\textbf{Comparison with the state-of-the-art. } The performances are evaluated on the Breakfast dataset as in Table~\ref{table:acti}. The performance of Bridge-Prompt fine-tuned features precedes I3D fine-tuned features. Since Bridge-Prompt is not a specially designed architecture for activity recognition, our straightforward prompt-based recognition approach may be inferior to more complicated recognition backbones based on fine-tuned I3D (\eg GHRM~\cite{zhou2021graph}). The performance can be further improved by combining Bridge-Prompt representations with other high-level backbones. 

\subsection{Ablation Studies}

We perform several ablation studies on the GTEA
dataset. Several adjustments have been conducted to evaluate
the influence of different settings.

\noindent\textbf{Fusion approaches. } We have studied more kinds of fusion strategies to integrate statistical or ordinal information into frame-wise features. They are listed as follows:

\begin{table}[]
\small
\begin{center}
\setlength{\tabcolsep}{5pt}
\caption{Comparisons of different fusion strategies for Bridge-Prompt by action segmentation results on GTEA dataset (split \#1).}
\vspace{-5pt}
\begin{tabular}{l|ccc|c|c}
\toprule
\textbf{Fusion strategy} & \multicolumn{3}{c|}{\textbf{F1}@\{10,25,50\}} & \textbf{Edit} & \textbf{Acc} \\ \hline
(a)    Vision-only & 90.3 & 87.4 & 76.5 & 86.2 & 81.0 \\
(b)  Pos-embedding i. & 89.1 & 86.2 & 77.5 & 84.8 & 80.0 \\
(b) Pos-embedding ii. & 88.7 & 87.3 & 76.4 & 84.0 & 79.5 \\
(c)    Weights for avg. & \textbf{91.8} & 88.1 & 79.1 & 86.5 & \textbf{83.7} \\ \hline
\textbf{(d)    Early-fusion} & 91.0 & \textbf{89.6} & \textbf{82.1} & \textbf{88.7} & 81.2 \\
\bottomrule
\end{tabular}\label{table:fuse}
\vspace{-20pt}
\end{center}
\end{table}


\noindent\textbf{(a) Vision-only fusion.} Within the vision-only fusion, only the frame-wise features are regarded as inputs of the fusion Transformer. The output clip-wise features are contrastively learned together with statistical prompts, semantical prompts, and integrated prompts.

\noindent\textbf{(b) Ordinal prompt fused as positional embedding.} The ordinal prompt embedding can be linearly projected as an embedded vector with its length equal to the clip length. Then it is added to the input frame-wise features as part of the positional embedding after a mapping operation. There are two ways of mapping: i. repeating the embedded vector along the width dimension; ii. computing the outer product between the embedded vector and ordinal prompt embedding. The output clip-wise features are contrastively learned together with all formats of text prompts as (a).

\noindent\textbf{(c) Ordinal prompt fused as weights of average.} The ordinal prompt embedding can be linearly projected as a weight vector with its length equal to the clip length. Then it is served as the weights of pooling operation for the input frame-wise features. The output weights are punished by an L2 loss function to avoid acquiring impulse-shape weights. The output clip-wise features are contrastively learned together with all formats of text prompts as (a) and (b).

\noindent\textbf{(d) Early-fused ordinal prompts with a learnable count token.} This is the fusion strategy adopted in our framework.


The action segmentation performances of different fusion strategies for Bridge-Prompt are evaluated on GTEA (split \#1). Table~\ref{table:fuse} shows the quantitative results, which indicates that the fusion module is significant for improving the learning effectiveness of Bridge-Prompt. By merging ordinal information into the fusion module, the learned representations possess the focused information for each ordinal action. The fusion strategy (b) and (c) are more direct ways to integrate ordinal prompts, however, the ordinal prompt embeddings are not cross-attentioned with vision features.
Specifically, the strategy (b) and (c) learn the information like ``where may the first action be in any 16-frame video clip?", while (d) focuses on ``where is the first action among all the actions in this video?".
The location for each ordinal action also depends on other adjacent actions, which makes the early-fusion way more convincible.

\begin{table}[]
\small
\begin{center}
\setlength{\tabcolsep}{5pt}
\caption{Comparisons of different loss choices for Bridge-Prompt by action segmentation results on GTEA dataset (split \#1).}
\vspace{-5pt}
\begin{tabular}{l|ccc|c|c}
\toprule
\textbf{Loss components} & \multicolumn{3}{c|}{\textbf{F1@\{10,25,50\}}} & \textbf{Edit} & \textbf{Acc} \\ \hline
$\mathcal{L}_{sem}$ & 87.4 & 82.5 & 70.6 & 81.9 & 79.5 \\
$\mathcal{L}_{sem}$+$\mathcal{L}_{integ}$ & 88.6 & 83.6 & 77.1 & 83.3 & \textbf{81.2} \\ \hline
$\mathcal{L}_{sem}$+$\mathcal{L}_{integ}$+$\mathcal{L}_{stat}$ & \textbf{91.0} & \textbf{89.6} & \textbf{82.1} & \textbf{88.7} & \textbf{81.2} \\
\bottomrule
\end{tabular}\label{table:loss}
\vspace{-20pt}
\end{center}
\end{table}

\noindent\textbf{Choice for loss functions. } In our design, we consider three main components in the loss function: semantics, integrated semantics, and statistics. We perform ablation experiments to test the effectiveness of all three loss components. Table~\ref{table:loss} shows the quantitative results, which indicates that all three losses make positive contributions to the final performance. It is reasonable since all the three text components are combined to depict both contextual and out-of-context semantics for a series of ordinal actions.

\noindent\textbf{Transferability studies. } Text is a flexible and extensible form of supervision. Different from class IDs, knowledge in texts can be transferred to unseen forms of script based on the generalization ability of pre-trained language models. To verify the transferability of Bridge-Prompt, we conduct a test on the prompt-based ordinal action inferences. For humans, action knowledge can be transferred between similar activities. As an example, a person can possibly learn how to \textit{make tea} if he/she knows how to \textit{make coffee}, since the sub-actions of the two activities are highly similar. For a class ID-based model, it is unable to transfer the knowledge between similar activities without manual interventions. Under prompt-based inferences, it is as simple as replacing the filling-in words in prompts. To quantitatively explain the transfer effects, we conduct experiments by training the framework on one human activity and evaluating the prompt inference accuracy on another one. The results are displayed in Table~\ref{table:transf}, which indicate that Bridge-Prompt has a promising zero-shot transferability.


\begin{table}[]
\small
\begin{center}
\setlength{\tabcolsep}{3pt}
\caption{Prompt-based inference accuracies on GTEA. (coffee2tea refers to transferring the knowledge of \textit{making coffee} to \textit{making tea}, and so forth; AKL refers to training with all-knowing labels.)}
\vspace{-5pt}
\begin{tabular}{c|cccc|c}
\toprule
\multicolumn{1}{l|}{\textbf{Trans-type}}
& \begin{tabular}[c]{@{}c@{}}coffee2\\ tea\end{tabular} &
\begin{tabular}[c]{@{}c@{}}cofhoney2\\ tea\end{tabular} &\begin{tabular}[c]{@{}c@{}}hotdog2\\ pealate\end{tabular} & \begin{tabular}[c]{@{}c@{}}peanut2\\ pealate\end{tabular} & \begin{tabular}[c]{@{}c@{}}overall\\(AKL)\end{tabular} \\ \hline
\textbf{top-1 Acc(\%)} & 38.8 & 41.7 & 15.5 & 24.6 & 54.5 \\
\textbf{top-5 Acc(\%)} & 74.4 & 81.3 & 45.1 & 54.8 & 90.9 \\ \bottomrule
\end{tabular}\label{table:transf}
\vspace{-20pt}
\end{center}
\end{table}





\section{Conclusion and Discussion}

In this paper, we have focused on the issue of ordinal action analysis in instructional videos. We proposed a prompt-based learning framework, Bridge-Prompt, which models the semantic relations across ordinal actions. To capture both out-of-context and contextual information of ordinal actions, text prompts are designed to integrate statistical, ordinal, and semantic information. Further experiments are conducted on two downstream tasks including action segmentation and long-term action recognition. The results have demonstrated that Bridge-Prompt has strong capability in the analysis of ordinal actions. 

\noindent\textbf{Limitations.} Language can abstract the semantics from raw tedious videos. Although it is appealing to conduct large-scale vision-language pre-training on massive instructional video datasets such as HowTo100M~\cite{miech2019howto100m}, we are limited by the computing resources. Fortunately, we find that the manual label is a more accurate and concise form of semantic abstraction. With the help of pre-trained language models, we are able to learn the semantics of ordinal actions in a more efficient and accurate way based on text supervision.

\noindent\textbf{Social impact.} Despite the adaptiveness and convenience of the prompt-based approach to collaborate with vision models, it also means that fake labels are easier to create. To protect the vision-language model from potential attacks, label-filtering mechanisms and model self-inspections should be considered in practical applications. 

\vspace{10pt}
\noindent\textbf{Acknowledgments   }
This work was supported in part by the National Key Research and Development Program of China under Grant 2017YFA0700802, in part by the National Natural Science Foundation of China under Grant 62125603 and Grant U1813218, in part by a grant from the Beijing Academy of Artificial Intelligence (BAAI), and in part by the Postdoctoral Innovative Talent Support Program of China under Grant BX2021160.

{\small
\bibliographystyle{ieee_fullname}
\bibliography{egbib}

\begin{thebibliography}{10}\itemsep=-1pt

\bibitem{ahn2021refining}
Hyemin Ahn and Dongheui Lee.
\newblock Refining action segmentation with hierarchical video representations.
\newblock In {\em ICCV}, pages 16302--16310, 2021.

\bibitem{apostolidis2021video}
Evlampios~E. Apostolidis, Eleni Adamantidou, Alexandros~I. Metsai, Vasileios
  Mezaris, and Ioannis Patras.
\newblock Video summarization using deep neural networks: {A} survey.
\newblock {\em Proc. {IEEE}}, 109(11):1838--1863, 2021.

\bibitem{Arnab_2021_ICCV}
Anurag Arnab, Mostafa Dehghani, Georg Heigold, Chen Sun, Mario Lu\v{c}i\'c, and
  Cordelia Schmid.
\newblock Vivit: A video vision transformer.
\newblock In {\em ICCV}, pages 6836--6846, 2021.

\bibitem{NEURIPS2020_1457c0d6}
Tom Brown, Benjamin Mann, Nick Ryder, Melanie Subbiah, Jared~D Kaplan, Prafulla
  Dhariwal, Arvind Neelakantan, Pranav Shyam, Girish Sastry, Amanda Askell,
  Sandhini Agarwal, Ariel Herbert-Voss, Gretchen Krueger, Tom Henighan, Rewon
  Child, Aditya Ramesh, Daniel Ziegler, Jeffrey Wu, Clemens Winter, Chris
  Hesse, Mark Chen, Eric Sigler, Mateusz Litwin, Scott Gray, Benjamin Chess,
  Jack Clark, Christopher Berner, Sam McCandlish, Alec Radford, Ilya Sutskever,
  and Dario Amodei.
\newblock Language models are few-shot learners.
\newblock In {\em NeurIPS}, pages 1877--1901, 2020.

\bibitem{carreira2017quo}
Joao Carreira and Andrew Zisserman.
\newblock Quo vadis, action recognition? a new model and the kinetics dataset.
\newblock In {\em CVPR}, pages 6299--6308, 2017.

\bibitem{chen2020action}
Min-Hung Chen, Baopu Li, Yingze Bao, Ghassan AlRegib, and Zsolt Kira.
\newblock Action segmentation with joint self-supervised temporal domain
  adaptation.
\newblock In {\em CVPR}, pages 9454--9463, 2020.

\bibitem{vit2021}
Alexey Dosovitskiy, Lucas Beyer, Alexander Kolesnikov, Dirk Weissenborn,
  Xiaohua Zhai, Thomas Unterthiner, Mostafa Dehghani, Matthias Minderer, Georg
  Heigold, Sylvain Gelly, Jakob Uszkoreit, and Neil Houlsby.
\newblock An image is worth 16x16 words: Transformers for image recognition at
  scale.
\newblock In {\em ICLR}, 2021.

\bibitem{Dzabraev_2021_CVPR}
Maksim Dzabraev, Maksim Kalashnikov, Stepan Komkov, and Aleksandr Petiushko.
\newblock Mdmmt: Multidomain multimodal transformer for video retrieval.
\newblock In {\em CVPRW}, pages 3354--3363, 2021.

\bibitem{farha2019ms}
Yazan~Abu Farha and Jurgen Gall.
\newblock Ms-tcn: Multi-stage temporal convolutional network for action
  segmentation.
\newblock In {\em ICCV}, pages 3575--3584, 2019.

\bibitem{fathi2011learning}
Alireza Fathi, Xiaofeng Ren, and James~M Rehg.
\newblock Learning to recognize objects in egocentric activities.
\newblock In {\em CVPR}, pages 3281--3288, 2011.

\bibitem{feichtenhofer2019slowfast}
Christoph Feichtenhofer, Haoqi Fan, Jitendra Malik, and Kaiming He.
\newblock Slowfast networks for video recognition.
\newblock In {\em CVPR}, pages 6202--6211, 2019.

\bibitem{gao2021clip}
Peng Gao, Shijie Geng, Renrui Zhang, Teli Ma, Rongyao Fang, Yongfeng Zhang,
  Hongsheng Li, and Yu Qiao.
\newblock Clip-adapter: Better vision-language models with feature adapters.
\newblock {\em arXiv preprint arXiv:2110.04544}, 2021.

\bibitem{gao2021global2local}
Shang-Hua Gao, Qi Han, Zhong-Yu Li, Pai Peng, Liang Wang, and Ming-Ming Cheng.
\newblock Global2local: Efficient structure search for video action
  segmentation.
\newblock In {\em CVPR}, pages 16805--16814, 2021.

\bibitem{girdhar2017actionvlad}
Rohit Girdhar, Deva Ramanan, Abhinav Gupta, Josef Sivic, and Bryan Russell.
\newblock Actionvlad: Learning spatio-temporal aggregation for action
  classification.
\newblock In {\em CVPR}, pages 971--980, 2017.

\bibitem{huang2020improving}
Yifei Huang, Yusuke Sugano, and Yoichi Sato.
\newblock Improving action segmentation via graph-based temporal reasoning.
\newblock In {\em CVPR}, pages 14024--14034, 2020.

\bibitem{hussein2019timeception}
Noureldien Hussein, Efstratios Gavves, and Arnold~WM Smeulders.
\newblock Timeception for complex action recognition.
\newblock In {\em CVPR}, pages 254--263, 2019.

\bibitem{hussein2019videograph}
Noureldien Hussein, Efstratios Gavves, and Arnold~WM Smeulders.
\newblock Videograph: Recognizing minutes-long human activities in videos.
\newblock In {\em ICCVW}, 2019.

\bibitem{ishikawa2021alleviating}
Yuchi Ishikawa, Seito Kasai, Yoshimitsu Aoki, and Hirokatsu Kataoka.
\newblock Alleviating over-segmentation errors by detecting action boundaries.
\newblock In {\em WACV}, pages 2322--2331, 2021.

\bibitem{align2021}
Chao Jia, Yinfei Yang, Ye Xia, Yi{-}Ting Chen, Zarana Parekh, Hieu Pham,
  Quoc~V. Le, Yun{-}Hsuan Sung, Zhen Li, and Tom Duerig.
\newblock Scaling up visual and vision-language representation learning with
  noisy text supervision.
\newblock In {\em ICML}, pages 4904--4916, 2021.

\bibitem{graph2017}
Thomas~N. Kipf and Max Welling.
\newblock Semi-supervised classification with graph convolutional networks.
\newblock In {\em ICLR}, 2017.

\bibitem{breakfast2014}
Hilde Kuehne, Ali~Bilgin Arslan, and Thomas Serre.
\newblock The language of actions: Recovering the syntax and semantics of
  goal-directed human activities.
\newblock In {\em CVPR}, pages 780--787, 2014.

\bibitem{lea2017temporal}
Colin Lea, Michael~D Flynn, Rene Vidal, Austin Reiter, and Gregory~D Hager.
\newblock Temporal convolutional networks for action segmentation and
  detection.
\newblock In {\em CVPR}, pages 156--165, 2017.

\bibitem{lea2016segmental}
Colin Lea, Austin Reiter, Ren{\'e} Vidal, and Gregory~D Hager.
\newblock Segmental spatiotemporal cnns for fine-grained action segmentation.
\newblock In {\em ECCV}, pages 36--52, 2016.

\bibitem{li2020ms}
Shi-Jie Li, Yazan AbuFarha, Yun Liu, Ming-Ming Cheng, and Juergen Gall.
\newblock Ms-tcn++: Multi-stage temporal convolutional network for action
  segmentation.
\newblock {\em IEEE TPAMI}, pages 1--1, 2020.

\bibitem{liu2021pre}
Pengfei Liu, Weizhe Yuan, Jinlan Fu, Zhengbao Jiang, Hiroaki Hayashi, and
  Graham Neubig.
\newblock Pre-train, prompt, and predict: A systematic survey of prompting
  methods in natural language processing.
\newblock {\em arXiv preprint arXiv:2107.13586}, 2021.

\bibitem{liu2021video}
Ze Liu, Jia Ning, Yue Cao, Yixuan Wei, Zheng Zhang, Stephen Lin, and Han Hu.
\newblock Video swin transformer.
\newblock {\em arXiv preprint arXiv:2106.13230}, 2021.

\bibitem{AdamW}
Ilya Loshchilov and Frank Hutter.
\newblock Decoupled weight decay regularization.
\newblock In {\em ICLR}, 2019.

\bibitem{Luo2020UniVL}
Huaishao Luo, Lei Ji, Botian Shi, Haoyang Huang, Nan Duan, Tianrui Li, Jason
  Li, Taroon Bharti, and Ming Zhou.
\newblock Univl: A unified video and language pre-training model for multimodal
  understanding and generation.
\newblock {\em arXiv preprint arXiv:2002.06353}, 2020.

\bibitem{miech2019howto100m}
Antoine Miech, Dimitri Zhukov, Jean-Baptiste Alayrac, Makarand Tapaswi, Ivan
  Laptev, and Josef Sivic.
\newblock Howto100m: Learning a text-video embedding by watching hundred
  million narrated video clips.
\newblock In {\em ICCV}, pages 2630--2640, 2019.

\bibitem{nawhal2021activity}
Megha Nawhal and Greg Mori.
\newblock Activity graph transformer for temporal action localization.
\newblock {\em arXiv preprint arXiv:2101.08540}, 2021.

\bibitem{CLIP2021}
Alec Radford, Jong~Wook Kim, Chris Hallacy, Aditya Ramesh, Gabriel Goh,
  Sandhini Agarwal, Girish Sastry, Amanda Askell, Pamela Mishkin, Jack Clark,
  Gretchen Krueger, and Ilya Sutskever.
\newblock Learning transferable visual models from natural language
  supervision.
\newblock In {\em ICML}, pages 8748--8763, 2021.

\bibitem{PET2021}
Timo Schick and Hinrich Sch{\"{u}}tze.
\newblock Exploiting cloze-questions for few-shot text classification and
  natural language inference.
\newblock In {\em EACL}, pages 255--269, 2021.

\bibitem{shao2020intra}
Dian Shao, Yue Zhao, Bo Dai, and Dahua Lin.
\newblock Intra-and inter-action understanding via temporal action parsing.
\newblock In {\em CVPR}, pages 730--739, 2020.

\bibitem{autoprompt:emnlp20}
Taylor Shin, Yasaman Razeghi, Robert L.~Logan IV, Eric Wallace, and Sameer
  Singh.
\newblock {AutoPrompt}: Eliciting knowledge from language models with
  automatically generated prompts.
\newblock In {\em EMNLP}, pages 4222--4235, 2020.

\bibitem{2014stream}
Karen Simonyan and Andrew Zisserman.
\newblock Two-stream convolutional networks for action recognition in videos.
\newblock In {\em NeurIPS}, pages 568--576, 2014.

\bibitem{soomro2012ucf101}
Khurram Soomro, Amir~Roshan Zamir, and Mubarak Shah.
\newblock Ucf101: A dataset of 101 human actions classes from videos in the
  wild.
\newblock {\em arXiv preprint arXiv:1212.0402}, 2012.

\bibitem{stein2013combining}
Sebastian Stein and Stephen~J McKenna.
\newblock Combining embedded accelerometers with computer vision for
  recognizing food preparation activities.
\newblock In {\em ACM}, pages 729--738, 2013.

\bibitem{tang2019coin}
Yansong Tang, Dajun Ding, Yongming Rao, Yu Zheng, Danyang Zhang, Lili Zhao,
  Jiwen Lu, and Jie Zhou.
\newblock Coin: A large-scale dataset for comprehensive instructional video
  analysis.
\newblock In {\em CVPR}, pages 1207--1216, 2019.

\bibitem{3dcnn}
Du Tran, Lubomir~D. Bourdev, Rob Fergus, Lorenzo Torresani, and Manohar Paluri.
\newblock Learning spatiotemporal features with 3d convolutional networks.
\newblock In {\em ICCV}, pages 4489--4497, 2015.

\bibitem{wang2021actionclip}
Mengmeng Wang, Jiazheng Xing, and Yong Liu.
\newblock Actionclip: A new paradigm for video action recognition.
\newblock {\em arXiv preprint arXiv:2109.08472}, 2021.

\bibitem{wang2020boundary}
Zhenzhi Wang, Ziteng Gao, Limin Wang, Zhifeng Li, and Gangshan Wu.
\newblock Boundary-aware cascade networks for temporal action segmentation.
\newblock In {\em ECCV}, pages 34--51, 2020.

\bibitem{yi2021asformer}
Fangqiu Yi, Hongyu Wen, and Tingting Jiang.
\newblock Asformer: Transformer for action segmentation.
\newblock {\em arXiv preprint arXiv:2110.08568}, 2021.

\bibitem{zhou2021graph}
Jiaming Zhou, Kun-Yu Lin, Haoxin Li, and Wei-Shi Zheng.
\newblock Graph-based high-order relation modeling for long-term action
  recognition.
\newblock In {\em ICCV}, pages 8984--8993, 2021.

\bibitem{zhou2021coop}
Kaiyang Zhou, Jingkang Yang, Chen~Change Loy, and Ziwei Liu.
\newblock Learning to prompt for vision-language models.
\newblock {\em arXiv preprint arXiv:2109.01134}, 2021.

\bibitem{zhou2018towards}
Luowei Zhou, Chenliang Xu, and Jason~J Corso.
\newblock Towards automatic learning of procedures from web instructional
  videos.
\newblock In {\em AAAI}, pages 7590--7598, 2018.

\bibitem{2019Cross}
D Zhukov, J.~B. Alayrac, R.~G. Cinbis, D Fouhey, I. Laptev, and J. Sivic.
\newblock Cross-task weakly supervised learning from instructional videos.
\newblock In {\em CVPR}, pages 3537--3545, 2019.

\bibitem{zhukov2019cross}
Dimitri Zhukov, Jean-Baptiste Alayrac, Ramazan~Gokberk Cinbis, David Fouhey,
  Ivan Laptev, and Josef Sivic.
\newblock Cross-task weakly supervised learning from instructional videos.
\newblock In {\em CVPR}, pages 3537--3545, 2019.

\end{thebibliography}
}

\end{document}